\documentclass[conference]{IEEEtran}
\IEEEoverridecommandlockouts
\usepackage{comment}
\usepackage{graphicx}
\usepackage{multirow}
\usepackage{cite}
\usepackage{longtable}
\usepackage{xurl}
\usepackage{amssymb}
\usepackage{float}
\usepackage{array}
\newcolumntype{P}[1]{>{\centering\arraybackslash}p{#1}}
\def\BibTeX{{\rm B\kern-.05em{\sc i\kern-.025em b}\kern-.08em
    T\kern-.1667em\lower.7ex\hbox{E}\kern-.125emX}}
\begin{document}

\title{Impact of Attention on Adversarial Robustness of Image Classification Models\\
}

\author{\IEEEauthorblockN{Prachi Agrawal}
\IEEEauthorblockA{\textit{Dept. of Computer Science and Information Systems} \\
\textit{Birla Institute of Technology and Science}\\
Pilani, India \\
agr.prachi22@gmail.com}
\\
\IEEEauthorblockN{Sanjay Kumar Sonbhadra}
\IEEEauthorblockA{\textit{Dept. of Information Technology} \\
\textit{Indian Institute of Information Technology Allahabad}\\
Prayagraj, India \\
rsi2017502@iiita.ac.in}
\and
\IEEEauthorblockN{Narinder Singh Punn}
\IEEEauthorblockA{\textit{Dept. of Information Technology} \\
\textit{Indian Institute of Information Technology Allahabad}\\
Prayagraj, India \\
pse2017002@iiita.ac.in}
\\
\IEEEauthorblockN{Sonali Agarwal}
\IEEEauthorblockA{\textit{Dept. of Information Technology} \\
\textit{Indian Institute of Information Technology Allahabad}\\
Prayagraj, India \\
sonali@iiita.ac.in}
}

\maketitle

\begin{abstract}
Adversarial attacks against deep learning models have gained significant attention and recent works have proposed explanations for the existence of adversarial examples and techniques to defend the models against these attacks. Attention in computer vision has been used to incorporate focused learning of important features and has led to improved accuracy. Recently, models with attention mechanisms have been proposed to enhance adversarial robustness. Following this context, this work aims at a general understanding of the impact of attention on adversarial robustness. This work presents a comparative study of adversarial robustness of non-attention and attention based image classification models trained on CIFAR-10, CIFAR-100 and Fashion MNIST datasets under the popular white box and black box attacks. The experimental results show that the robustness of attention based models may be dependent on the datasets used i.e. the number of classes involved in the classification. In contrast to the datasets with less number of classes, attention based models are observed to show better robustness towards classification.
\end{abstract}


\section{Introduction}
Recently, the vulnerability of deep learning models to adversarial attacks has been recognized. Adversarial attacks in the context of image classification involve fooling an image classifier by generating an image with small perturbation from the original image which results in misclassification of the image. Previous works in the field have proposed a variety of explanations for the existence of adversarial examples including high-dimensional geometry of data manifold~\cite{gilmer2018adversarial} and computational limitations of learning algorithms~\cite{bubeck2019adversarial}. Ilyas et al.~\cite{ilyas2019adversarial} emphasize that adversarial robustness is a human-centric phenomenon and vulnerability to adversarial examples exists because models learn to treat non-robust features and robust features with equal importance. 

Attention plays a huge role in the human visual system. Humans use a sequence of partial glimpses and selectively focus on important parts instead of processing the entire scene at once. Inspired by the attention mechanism in the human visual system, several methods have been proposed to incorporate attention in convolution neural network (CNN) based image classification tasks which have also led to improvement in the accuracy of the classifiers e.g. residual attention network~\cite{wang2017residual}, squeeze and excitation networks~\cite{hu2018squeeze}, convolutional block attention module~\cite{woo2018cbam}, etc.~\cite{punn2021modality}.

Following this context, the present work focuses on studying the impact of learning salient features through attention mechanisms on the adversarial robustness of a model using the image classification problem. The major contribution of the present research work is as follows:
\begin{itemize}
    \item Comparative analysis of the robustness of attention based (using mixed attention) and non-attention based models under the white box and transfer based black box attacks.
    \item Exploring the dependence of adversarial robustness of attention and non-attention based models on datasets used.
    \item While recent works have proposed attention based techniques to improve adversarial robustness, our study focuses on understanding how attention impacts robustness which would give more insights on the existence of adversarial examples and how defense strategies could be developed.
    \item To the best of our knowledge, this is also the first attempt to analyze the adversarial robustness of image classification models with mixed attention.  

\end{itemize}

The rest paper is divided into several sections, where Section 2 describes the literature survey of the works describing robustness of the classification models, followed by Section 3 presenting the proposed framework to simulate white box and black box attacks on classification models with different datasets. Section 4 presents the the simulation results of the attacks on the models. Finally, concluding remarks are presented in Section 5.

\section{Related work}
In order to defend against adversarial attacks, extensive research has been conducted to develop defense strategies and build robust models. Some of them are robust training, input transformation, randomization, model ensemble, and certified defenses. 

In robust training, the classifier is made robust against small noises internally. Adversarial training~\cite{goodfellow2014explaining, madry2017towards, tramer2017ensemble} is a popular technique based on robust training which augments training data by adversarial examples. Several methods based on robust training using losses or regularizations have been proposed. These include regularization on input gradient~\cite{ross2018improving}, perturbation based regularization~\cite{he2016deep} and variants on network Lipschitz constant~\cite{cisse2017parseval}. Another technique is to transform the inputs before feeding them into the classifier. Total variance minimization~\cite{guo2017countering}, autoencoder based denoising~\cite{liao2018defense}, using generative models to project adversarial examples onto the data distribution~\cite{samangouei2018defense} are some of the input transformation based strategies. Adding randomness to the input~\cite{xie2017mitigating, pang2019mixup} or the model~\cite{liu2018towards} have also been proposed to enhance robustness. Another defense strategy is to construct an ensemble of models and aggregate the output of each model in the ensemble~\cite{kurakin2018adversarial, pang2019improving}. There are a lot of works based on certified defenses which are guaranteed to be robust against adversarial perturbations under some attack models~\cite{raghunathan2018certified, sinha2017certifying, xiao2018training, wong2018scaling}.

Several recent works have proposed models with attention based mechanisms to enhance robustness. Zoran et al.~\cite{zoran2020towards} use a visual attention component guided by a recurrent top down approach to improve the adversarial robustness, whereas Liu et al.~\cite{liu2018feature} propose a nonlinear attention module for feature prioritization to improve adversarial robustness. In similar work, Zhao et al.~\cite{zhao2020exploring} show that self-attention networks may have benefits in terms of adversarial robustness and later, Vaishnavi et al.~\cite{vaishnavi2020can} use foreground attention masks i.e separate the background from the foreground and mask the background before training and classification. Their results show an increase in adversarial robustness on using foreground masks. Goodman et al.~\cite{goodman2019improving} develop a method called attention and adversarial logit pairing that encourages both attention map and logit for pairs of clean examples and their adversarial counterparts to be similar. In this work, we attempt to get a general understanding of the impact of attention in image classification models using mixed attention under different kinds of attacks.

\section{Proposed work}
\begin{figure}[]
    \centering
    \includegraphics[width=\linewidth]{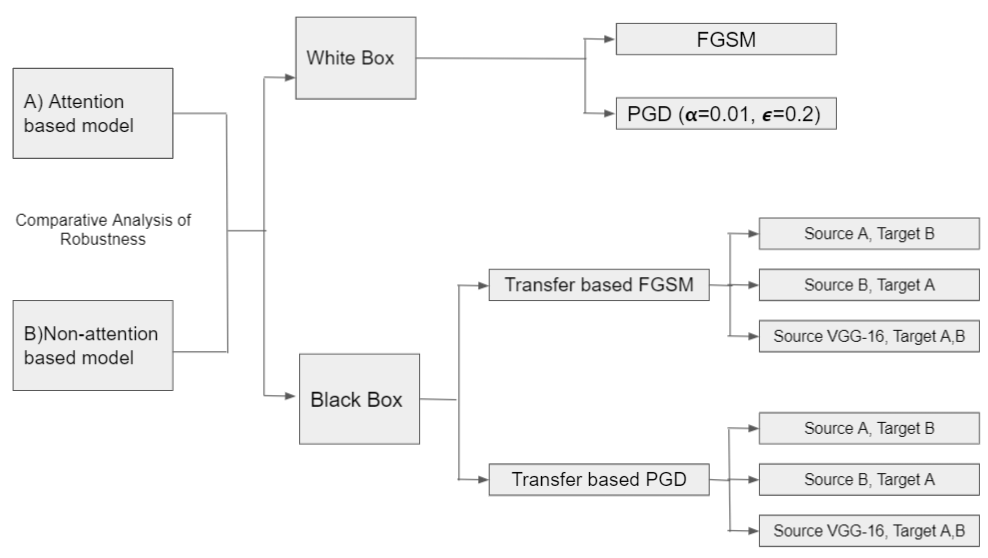}
    \caption{Simulation framework for the attacks on attention and non-attention classification models.}
    \label{fig1}
\end{figure}

Based on the discussion in Section 1, we conduct experiments to understand if learning of salient features through attention mechanisms translates to adversarial robustness through better identification of underlying features in adversarial images. We compare the adversarial robustness of an image classifier with and without attention under the white box and black box attacks. Fig~\ref{fig1} shows a schematic representation of the proposed framework in which the attacks are simulated.

We have considered Resnet-50~\cite{ross2018improving} for image classification and used Convolutional Block Attention Module (CBAM) to incorporate attention. CBAM contains two sequential sub-modules i.e. channel attention module which feature maps are important and spatial attention module conveys what within the feature map is important. A comparison of robustness of Resnet-50 and Resnet-50 with CBAM modules (i.e. CBAM+Resnet-50) is performed under different attacks.

\subsection{Attacks}
The attacks used in this work can be classified into white box and black box. In white box attacks, an attacker has complete access to the target model including the model parameters, architecture, training method and training data, and exploits this information to craft an attack. In black box attacks, the attacker has no access to model parameters and uses a different model or a query based strategy to perform an attack.

\subsubsection{White box attack}
Popular white box attacks include fast gradient sign method \cite{goodfellow2014explaining} (FGSM), projected gradient descent~\cite{madry2017towards} (PGD), deepfool~\cite{moosavi2016deepfool}, C\&W~\cite{carlini2017towards}. In this work, FGSM and PGD have been used to evaluate and compare the performance of attention and non-attention models under white box attack scenarios. 

FGSM is a single step gradient based approach where an adversarial example is created by maximizing the loss function $\mathcal{L}(x_{adv},y)$ wrt to the input $x$ and then adding the sign of output gradient to $x$ in order to produce $x_{adv}$ (adversarial example) and is represented in Eq.~\ref{eq1}. In our analysis, the accuracy of the target models is studied for various values of epsilon in the range of 0 to 0.3. PGD uses the projected gradient descent to iteratively create adversarial examples. The perturbed data in step $t+1$ can be represented in Eq.~\ref{eq2}.

\begin{equation}
    x_{adv}=x+\epsilon*sign(\triangledown_x \mathcal{L}(x,y))
    \label{eq1}
\end{equation}
\begin{equation}
    x^{t+1}_{adv}=\pi_{x+S}(x^{t}_{adv}+\epsilon+\alpha(sign(\triangledown_x \mathcal{L}(x^{t}_{adv},y))
    \label{eq2}
\end{equation}
Here, $\alpha$ is the step size and $\pi_{x+S}$ denotes projecting perturbations into set $S$. In all our experiments, we use PGD bounded under $l_{inf}$ norm with $\epsilon=0.2$ and $\alpha=0.01$. The attack accuracy is evaluated for different numbers of iterations.

\subsubsection{Black box attack}
Transfer based black box attack utilizes cross-model transferability of adversarial samples, which trains a local substitute model (source model) to generate adversarial examples and uses these examples to attack the target model. In this work, we use FGSM and PGD ($\epsilon$=0.2, $\alpha$=0.01) to generate examples from substitute model and carry out the transfer based black box attack in the following scenarios:
\begin{itemize}
    \item Non-attention model as the source and attention based model as target.
    \item Attention based model as the source and non-attention model as target.
    \item Comparison of attack against attention based model and non-attention model with VGG-16 as source.
\end{itemize}
The first scenario is used to understand if attention module helps in building defense and thus better robustness towards adversarial examples generated by non-attention based model. In second scenario, we evaluate if the attack on non-attention model using images generated by attention based model is better due to the learning of salient features by attention models. The final scenario helps in comparison of the robustness of attention and non-attention model in transfer based attack using a common model as source i.e. VGG16. The attacks setup of the present research work is presented in Table~\ref{tab1} for classification models over multiple datasets.

\begin{table}[]
\centering
\caption{Summary of attacks used to check the robustness of the classification models.}
\label{tab1}
\begin{tabular}{|p{2in}|p{1in}|}
\hline
\textbf{Attack}                                                                                                                     & \textbf{Type}              \\ \hline
Fast gradient sign method (FGSM)                                                                                                                                & White Box (Gradient Based) \\ \hline
Projected gradient descent (PGD)                                                                                                                                 & White Box (Gradient Based) \\ \hline
Transfer based FGSM \& PGD ; Source: Attention based model   (CBAM+Resnet-50) , Target: Non-attention model (Resnet-50)             & Black Box (Transfer based) \\ \hline
Transfer based FGSM \& PGD, Source: Non-attention model   (Resnet-50) , Target: Attention based model (CBAM+Resnet-50)              & Black Box (Transfer based) \\ \hline
Transfer based FGSM \& PGD , Source: VGG-16 , Target:   Attention model (CBAM+Resnet-50) \& non-attention based model   (Resnet-50) & Black Box (Transfer based) \\ \hline
\end{tabular}
\end{table}
\subsection{Datasets used}
Publicly available datasets: CIFAR-10 \cite{krizhevsky2009learning}, CIFAR-100 \cite{krizhevsky2009learning} and Fashion MNIST \cite{xiao2017fashion} have been used in our work. CIFAR-10 and CIFAR-100 consist of 60k colour images of size 32x32 the training set consisting of 50k images and the test set of 10k images. CIFAR-10 has 10 classes of 6k images each and CIFAR-100 has 100 classes of 600 images each. Fashion MNIST dataset consists of grayscale images of 10 fashion classes. The size of images is 28x28 and consists of 60k training images and 10k testing images.

\section{Results and discussion}
\subsection{White box}
Table~\ref{tab2} shows the performance of the non-attention model and attention based model trained on CIFAR-10, CIFAR-100 and FashionMNIST datasets under FGSM and PGD attacks with different values of perturbation and number of iterations respectively.
\begin{table*}[]
\centering
\caption{Summary of results of white box attacks.}
\label{tab2}
\begin{tabular}{|P{0.2in}|c||P{0.7in}|P{0.7in}||P{0.7in}|P{0.7in}||P{0.7in}|P{0.7in}|}
\hline
\multicolumn{2}{|c||}{\multirow{2}{*}{\textbf{Attacks}}}     & \multicolumn{6}{c|}{\textbf{Dataset}}                                                                                                                                          \\ \cline{3-8} 
\multicolumn{2}{|c||}{}                                      & \multicolumn{2}{c||}{\textbf{CIFAR-10}}                   & \multicolumn{2}{c||}{\textbf{CIFAR-100}}                  & \multicolumn{2}{c|}{\textbf{Fashion MNIST}}              \\ \hline
\multirow{10}{*}{\rotatebox[origin=c]{90}{\textbf{FGSM}}} & \textbf{Perturbation}      & \textbf{Resnet-50 (Acc)} & \textbf{CBAM+ Resnet-50 (Acc)} & \textbf{Resnet-50 (Acc)} & \textbf{CBAM+ Resnet-50 (Acc)} & \textbf{Resnet-50 (Acc)} & \textbf{CBAM+ Resnet-50 (Acc)} \\ \cline{2-8} 
                               & 0                          & 83.03                    & 80.4                          & 77.2                     & 75.69                         & 89.89                    & 92.73                         \\ \cline{2-8} 
                               & 0.05                       & 13.57                    & 0.37                          & 13.4                     & 19.29                         & 42.42                    & 17.59                         \\ \cline{2-8} 
                               & 0.1                        & 12.4                     & 1.79                          & 4.2                      & 5.43                          & 6.63                     & 6.85                          \\ \cline{2-8} 
                               & 0.15                       & 11.91                    & 4.6                           & 1.49                     & 2.43                          & 6.89                     & 6.55                          \\ \cline{2-8} 
                               & 0.2                        & 11.34                    & 5.95                          & 0.97                     & 1.39                          & 6.73                     & 6.89                          \\ \cline{2-8} 
                               & 0.25                       & 10.88                    & 6.51                          & 0.93                     & 1.12                          & 6.86                     & 7.03                          \\ \cline{2-8} 
                               & 0.3                        & 9.3                      & 6.85                          & 0.92                     & 1.03                          & 7.16                     & 7.22                          \\ \hline \hline
\multirow{6}{*}{\rotatebox[origin=c]{90}{\textbf{PGD}}}  & \textbf{Iterations} & \textbf{Resnet-50 (Err)} & \textbf{CBAM+ Resnet-50 (Err)} & \textbf{Resnet-50 (Err)} & \textbf{CBAM+ Resnet-50 (Err)} & \textbf{Resnet-50 (Err)} & \textbf{CBAM+ Resnet-50 (Err)} \\ \cline{2-8} 
                               & 5                          & 99.98                    & 100                           & 99.83                    & 87.4                          & 95.68                    & 100                           \\ \cline{2-8} 
                               & 10                         & 100                      & 100                           & 100                      & 93.41                         & 100                      & 100                           \\ \cline{2-8} 
                               & 20                         & 100                      & 100                           & 100                      & 97.16                         & 100                      & 100                           \\ \hline
\end{tabular}
\end{table*}
\subsubsection{CIFAR-10}
\begin{figure}[H]
    \centering
    \includegraphics[width=\linewidth]{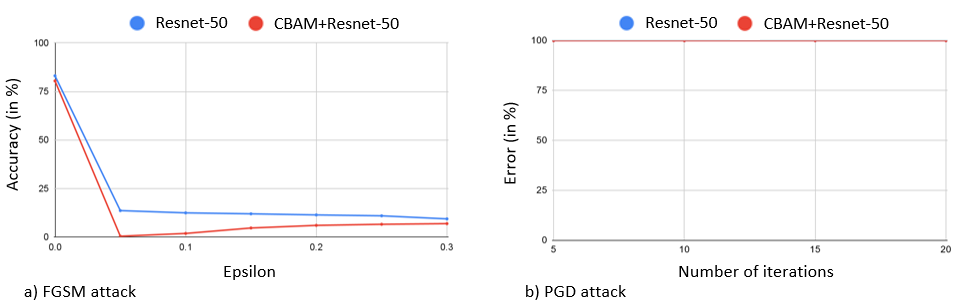}
    \caption{a) FGSM attack and b) PGD attack on Resnet-50 and CBAM+Resnet-50 for CIFAR-10 dataset.}
    \label{fig2}
\end{figure}

A comparison of white box FGSM attack for different perturbation values shown in Fig.~\ref{fig2}(a) suggests that CBAM+Resnet-50 is attacked to a much larger extent as compared to Resnet-50 throughout the range of epsilon values from 0.05 to 0.3, where the accuracy of CBAM+Resnet-50 even drops to $\sim0$\%. The white box PGD attack in Fig.~\ref{fig2}(b) also represents towards better robustness of the non-attention model (Resnet-50) since it takes a larger number of iterations as compared to the attention model (CBAM+Resnet-50) for complete attack i.e. 100\% error.

\subsubsection{CIFAR-100}
\begin{figure}[H]
    \centering
    \includegraphics[width=\linewidth]{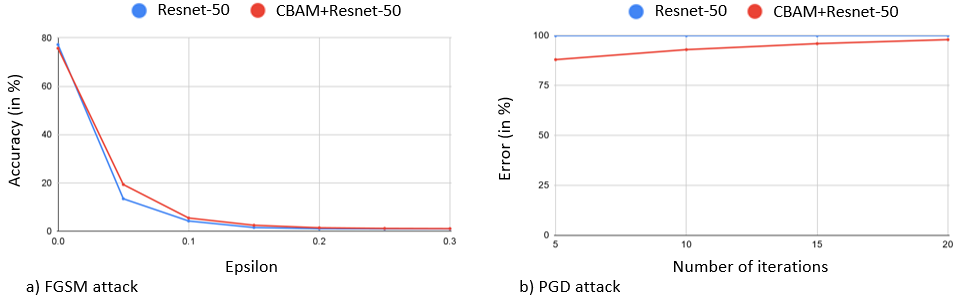}
    \caption{a) FGSM attack and b) PGD attack on Resnet-50 and CBAM+Resnet-50 for CIFAR-100 dataset.}
    \label{fig3}
\end{figure}

Unlike the observation for white box attacks on the CIFAR-10 dataset, the attention based model shows better robustness for the CIFAR-100 dataset. White box FGSM attack (Fig.~\ref{fig3}(a) shows slightly better accuracy for CBAM+Resnet-50 but the PGD attack (Fig.~\ref{fig3}(b) convincingly describes that CBAM+Resnet-50 is much more robust as compared to Resnet-50 since it isn’t attacked completely even in more than 20 iterations while Resnet-50 is completely attacked in less than 5 iterations.

\subsubsection{Fashion MNIST}
\begin{figure}[H]
    \centering
    \includegraphics[width=\linewidth]{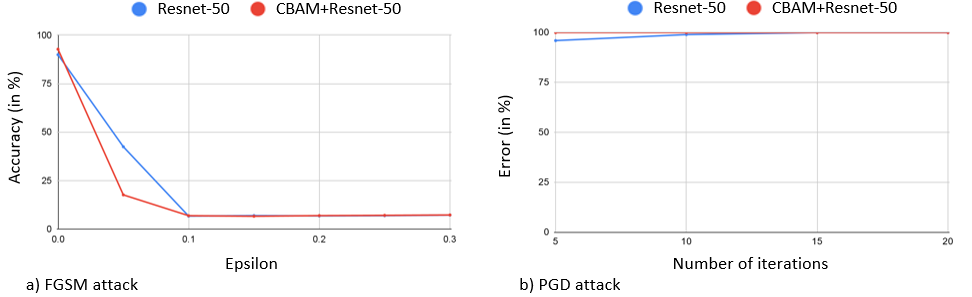}
    \caption{a) FGSM attack and b) PGD attack on Resnet-50 and CBAM+Resnet-50 for Fashion MNIST dataset.}
    \label{fig4}
\end{figure}

Fig.~\ref{fig4}(a) shows that Resnet-50 shows better adversarial robustness upto a perturbation value of 0.1. For epsilon values greater than 0.1, both CBAM+Resnet-50 and Resnet-50 have approximately equal accuracy under the FGSM attack. Fig.~\ref{fig4}(b) depicts the better robustness of Resnet-50 since it takes a larger number of iterations as compared to CBAM+Resnet-50 for complete attack.

\subsection{Black box}
Table~\ref{tab3} shows the performance of the non-attention model and attention based model trained on CIFAR-10, CIFAR-100 and FashionMNIST datasets under transfer based FGSM and PGD attacks under the 3 scenarios of source and target chosen which has been described in Table 1. The performance of the models have been observed for different values of perturbation in case of FGSM and different number of iterations for PGD.
\begin{table*}[!h]
\centering
\caption{Summary of results of black box attacks.}
\label{tab3}
\begin{tabular}{|c||P{0.2in}|c||P{0.5in}|P{0.5in}||P{0.5in}|P{0.5in}||P{0.5in}|P{0.5in}|P{0.5in}|}
\hline
\multicolumn{1}{|l||}{\textbf{Dataset}}  & \multicolumn{2}{c||}{\textbf{Attack}}                                        & \multicolumn{1}{P{0.5in}|}{\textbf{Source}}          & \multicolumn{1}{c||}{\textbf{Target}}               & \multicolumn{1}{c|}{\textbf{Source}}          & \multicolumn{1}{c||}{\textbf{Target}}               & \multicolumn{1}{c|}{\textbf{Source}}       & \multicolumn{1}{c|}{\textbf{Target-I}}        & \multicolumn{1}{c|}{\textbf{Target-II}}            \\ \hline
\multirow{15}{*}{\rotatebox[origin=c]{90}{\textbf{CIFAR-10}}}     & \multirow{10}{*}{\rotatebox[origin=c]{90}{\textbf{FGSM}}} & \multicolumn{1}{c||}{\textbf{Perturbation}} & \multicolumn{1}{P{0.5in}|}{\textbf{Resnet-50 (Acc)}} & \multicolumn{1}{P{0.5in}||}{\textbf{CBAM+ Resnet-50 (Acc)}} & \multicolumn{1}{P{0.5in}|}{\textbf{Resnet-50 (Acc)}} & \multicolumn{1}{P{0.5in}||}{\textbf{CBAM+ Resnet-50 (Acc)}} & \multicolumn{1}{P{0.5in}|}{\textbf{VGG-16 (Acc)}} & \multicolumn{1}{P{0.5in}|}{\textbf{Resnet-50 (Acc)}} & \multicolumn{1}{P{0.5in}|}{\textbf{CBAM+ Resnet-50 (Acc)}} \\ \cline{3-10} 
                                        &                                & 0                                          & 83.03                                         & 74.52                                              & 80.4                                          & 74.52                                              & 80.5                                       & 74.22                                         & 72.55                                              \\ \cline{3-10} 
                                        &                                & 0.05                                       & 13.57                                         & 37.6                                               & 0.37                                          & 42.54                                              & 5.93                                       & 25.57                                         & 28.07                                              \\ \cline{3-10} 
                                        &                                & 0.1                                        & 12.4                                          & 19                                                 & 1.79                                          & 20.91                                              & 6.64                                       & 13.82                                         & 14.33                                              \\ \cline{3-10} 
                                        &                                & 0.15                                       & 11.91                                         & 13.11                                              & 4.6                                           & 14.12                                              & 7.84                                       & 12.66                                         & 12.43                                              \\ \cline{3-10} 
                                        &                                & 0.2                                        & 11.34                                         & 11.65                                              & 5.95                                          & 11.83                                              & 8.66                                       & 12.89                                         & 11.2                                               \\ \cline{3-10} 
                                        &                                & 0.25                                       & 10.88                                         & 11.49                                              & 6.51                                          & 10.88                                              & 9.13                                       & 13.15                                         & 10.98                                              \\ \cline{3-10} 
                                        &                                & 0.3                                        & 9.3                                           & 11.58                                              & 6.85                                          & 10.47                                              & 9.47                                       & 8.84                                          & 10.69                                              \\ \cline{2-10} \cline{2-10}
                                        & \multirow{6}{*}{\rotatebox[origin=c]{90}{\textbf{PGD}}}  & \multicolumn{1}{c||}{\textbf{Iterations}}   & \multicolumn{1}{P{0.5in}|}{\textbf{Resnet-50 (Err)}} & \multicolumn{1}{P{0.5in}||}{\textbf{CBAM+ Resnet-50 (Err)}} & \multicolumn{1}{P{0.5in}|}{\textbf{Resnet-50 (Err)}} & \multicolumn{1}{P{0.5in}||}{\textbf{CBAM+ Resnet-50 (Err)}} & \multicolumn{1}{P{0.5in}|}{\textbf{VGG-16 (Err)}} & \multicolumn{1}{P{0.5in}|}{\textbf{Resnet-50 (Err)}} & \multicolumn{1}{P{0.5in}|}{\textbf{CBAM+ Resnet-50 (Err)}} \\ \cline{3-10} 
                                        &                                & 5                                          & 99.98                                         & 49.12                                              & 100                                           & 45.56                                              & 100                                        & 83.65                                         & 75.76                                              \\ \cline{3-10} 
                                        &                                & 10                                         & 100                                           & 64.47                                              & 100                                           & 59.03                                              & 100                                        & 95.51                                         & 92.44                                              \\ \cline{3-10} 
                                        &                                & 20                                         & 100                                           & 77.3                                               & 100                                           & 70.75                                              & 100                                        & 98.3                                          & 97.42                                              \\ \hline \hline
\multirow{15}{*}{\rotatebox[origin=c]{90}{\textbf{CIFAR-100}}}    & \multirow{10}{*}{\rotatebox[origin=c]{90}{\textbf{FGSM}}} & \multicolumn{1}{l||}{\textbf{Perturbation}} & \multicolumn{1}{P{0.5in}|}{\textbf{Resnet-50 (Acc)}} & \multicolumn{1}{P{0.5in}||}{\textbf{CBAM+ Resnet-50 (Acc)}} & \multicolumn{1}{P{0.5in}|}{\textbf{Resnet-50 (Acc)}} & \multicolumn{1}{P{0.5in}||}{\textbf{CBAM+ Resnet-50 (Acc)}} & \multicolumn{1}{P{0.5in}|}{\textbf{VGG-16 (Acc)}} & \multicolumn{1}{P{0.5in}|}{\textbf{Resnet-50 (Acc)}} & \multicolumn{1}{P{0.5in}|}{\textbf{CBAM+ Resnet-50 (Acc)}} \\ \cline{3-10} 
                                        &                                & 0                                          & 77.2                                          & 75.69                                              & 75.69                                         & 77.2                                               & 71.46                                      & 77.2                                          & 75.69                                              \\ \cline{3-10} 
                                        &                                & 0.05                                       & 13.4                                          & 18.3                                               & 19.29                                         & 21.3                                               & 5.37                                       & 16.23                                         & 15.74                                              \\ \cline{3-10} 
                                        &                                & 0.1                                        & 4.2                                           & 5.63                                               & 5.43                                          & 5.16                                               & 3.33                                       & 4.78                                          & 5.16                                               \\ \cline{3-10} 
                                        &                                & 0.15                                       & 1.49                                          & 2.21                                               & 2.43                                          & 1.58                                               & 2.21                                       & 1.72                                          & 2.18                                               \\ \cline{3-10} 
                                        &                                & 0.2                                        & 0.97                                          & 1.37                                               & 1.39                                          & 0.95                                               & 1.53                                       & 1.01                                          & 1.42                                               \\ \cline{3-10} 
                                        &                                & 0.25                                       & 0.93                                          & 1.16                                               & 1.12                                          & 0.99                                               & 1.11                                       & 0.88                                          & 1.1                                                \\ \cline{3-10} 
                                        &                                & 0.3                                        & 0.92                                          & 1.05                                               & 1.03                                          & 0.88                                               & 1.02                                       & 0.88                                          & 0.95                                               \\ \cline{2-10} \cline{2-10}
                                        & \multirow{6}{*}{\rotatebox[origin=c]{90}{\textbf{PGD}}}  & \multicolumn{1}{c||}{\textbf{Iterations}}   & \multicolumn{1}{P{0.5in}|}{\textbf{Resnet-50 (Err)}} & \multicolumn{1}{P{0.5in}||}{\textbf{CBAM+ Resnet-50 (Err)}} & \multicolumn{1}{P{0.5in}|}{\textbf{Resnet-50 (Err)}} & \multicolumn{1}{P{0.5in}||}{\textbf{CBAM+ Resnet-50 (Err)}} & \multicolumn{1}{P{0.5in}|}{\textbf{VGG-16 (Err)}} & \multicolumn{1}{P{0.5in}|}{\textbf{Resnet-50 (Err)}} & \multicolumn{1}{P{0.5in}|}{\textbf{CBAM+ Resnet-50 (Err)}} \\ \cline{3-10} 
                                        &                                & 5                                          & 99.83                                         & 84.64                                              & 87.4                                          & 78.93                                              & 94.15                                      & 71.74                                         & 71.54                                              \\ \cline{3-10} 
                                        &                                & 10                                         & 100                                           & 90.21                                              & 93.41                                         & 87.11                                              & 95.24                                      & 82.82                                         & 82.18                                              \\ \cline{3-10} 
                                        &                                & 20                                         & 100                                           & 93.18                                              & 97.16                                         & 92.4                                               & 95.99                                      & 89.72                                         & 88.51                                              \\ \hline \hline
\multirow{15}{*}{\rotatebox[origin=c]{90}{\textbf{Fashion MNIST}}} & \multirow{10}{*}{\rotatebox[origin=c]{90}{\textbf{FGSM}}} & \multicolumn{1}{l||}{\textbf{Perturbation}} & \multicolumn{1}{P{0.5in}|}{\textbf{Resnet-50 (Acc)}} & \multicolumn{1}{P{0.5in}||}{\textbf{CBAM+ Resnet-50 (Acc)}} & \multicolumn{1}{P{0.5in}|}{\textbf{Resnet-50 (Acc)}} & \multicolumn{1}{P{0.5in}||}{\textbf{CBAM+ Resnet-50 (Acc)}} & \multicolumn{1}{P{0.5in}|}{\textbf{VGG-16 (Acc)}} & \multicolumn{1}{P{0.5in}|}{\textbf{Resnet-50 (Acc)}} & \multicolumn{1}{P{0.5in}|}{\textbf{CBAM+ Resnet-50 (Acc)}} \\ \cline{3-10} 
                                        &                                & 0                                          & 89.89                                         & 92.73                                              & 92.73                                         & 89.89                                              & 91.93                                      & 89.89                                         & 92.73                                              \\ \cline{3-10} 
                                        &                                & 0.05                                       & 42.42                                         & 38.34                                              & 17.59                                         & 69.29                                              & 14.69                                      & 58.83                                         & 54.67                                              \\ \cline{3-10} 
                                        &                                & 0.1                                        & 6.63                                          & 14.93                                              & 6.85                                          & 15.38                                              & 6.04                                       & 17.54                                         & 25.03                                              \\ \cline{3-10} 
                                        &                                & 0.15                                       & 6.89                                          & 9.95                                               & 6.55                                          & 9.67                                               & 5.5                                        & 10.33                                         & 13.85                                              \\ \cline{3-10} 
                                        &                                & 0.2                                        & 6.73                                          & 9.15                                               & 6.89                                          & 7.99                                               & 5.64                                       & 8.26                                          & 10.67                                              \\ \cline{3-10} 
                                        &                                & 0.25                                       & 6.86                                          & 9.05                                               & 7.03                                          & 7.06                                               & 6.27                                       & 7.76                                          & 8.93                                               \\ \cline{3-10} 
                                        &                                & 0.3                                        & 7.16                                          & 9.02                                               & 7.22                                          & 6.69                                               & 6.95                                       & 7.27                                          & 7.98                                               \\ \cline{2-10} \cline{2-10}
                                        & \multirow{6}{*}{\rotatebox[origin=c]{90}{\textbf{PGD}}}  & \multicolumn{1}{c||}{\textbf{Iterations}}   & \multicolumn{1}{P{0.5in}|}{\textbf{Resnet-50 (Err)}} & \multicolumn{1}{P{0.5in}||}{\textbf{CBAM+ Resnet-50 (Err)}} & \multicolumn{1}{P{0.5in}|}{\textbf{Resnet-50 (Err)}} & \multicolumn{1}{P{0.5in}||}{\textbf{CBAM+ Resnet-50 (Err)}} & \multicolumn{1}{P{0.5in}|}{\textbf{VGG-16 (Err)}} & \multicolumn{1}{P{0.5in}|}{\textbf{Resnet-50 (Err)}} & \multicolumn{1}{P{0.5in}|}{\textbf{CBAM+ Resnet-50 (Err)}} \\ \cline{3-10} 
                                        &                                & 5                                          & 95.68                                         & 73.99                                              & 100                                           & 35.33                                              & 90.51                                      & 37.63                                         & 60.22                                              \\ \cline{3-10} 
                                        &                                & 10                                         & 100                                           & 92.5                                               & 100                                           & 64.02                                              & 90.55                                      & 57.55                                         & 70.5                                               \\ \cline{3-10} 
                                        &                                & 20                                         & 100                                           & 97.93                                              & 100                                           & 92.21                                              & 90.55                                      & 71.42                                         & 79.66                                              \\ \hline
\end{tabular}
\end{table*}

\begin{figure}[h!]
    \centering
    \includegraphics[width=\linewidth]{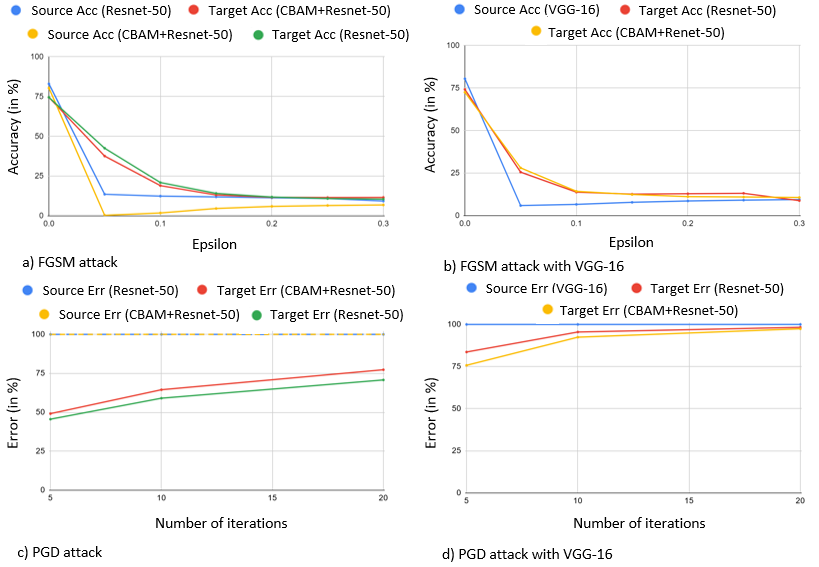}
    \caption{a) Transfer based FGSM, b) Transfer based FGSM with VGG-16 as source, c) Transfer based PGD and d) Transfer based PGD with VGG-16 as source for CIFAR-10 dataset.}
    \label{fig5}
\end{figure}

\subsubsection{CIFAR10}
For the FGSM attack, Resnet-50 showed better transferability leading to a better attack on CBAM+Resnet-50 as compared to an attack on Resnet-50 by examples generated by CBAM+Resnet-50 (Fig.~\ref{fig5}(a)). Under the FGSM transfer attack with VGG-16 as the source, the attention and non-attention model both showed approximately equal accuracy under attack (Fig.~\ref{fig5}(b)). Under the transfer based PGD attack, Resnet-50 was comparatively more robust to attack against adversarial examples generated by CBAM+Resnet-50 (Fig.~\ref{fig5}(c)). Resnet-50 thus showed better transferability for the attack against attention based model. However, under transfer based PGD with VGG-16 as the source, CBAM+Resnet-50 shows better robustness as compared to Resnet-50 (Fig.~\ref{fig5}(d)).

\begin{figure}[b!]
    \centering
    \includegraphics[width=\linewidth]{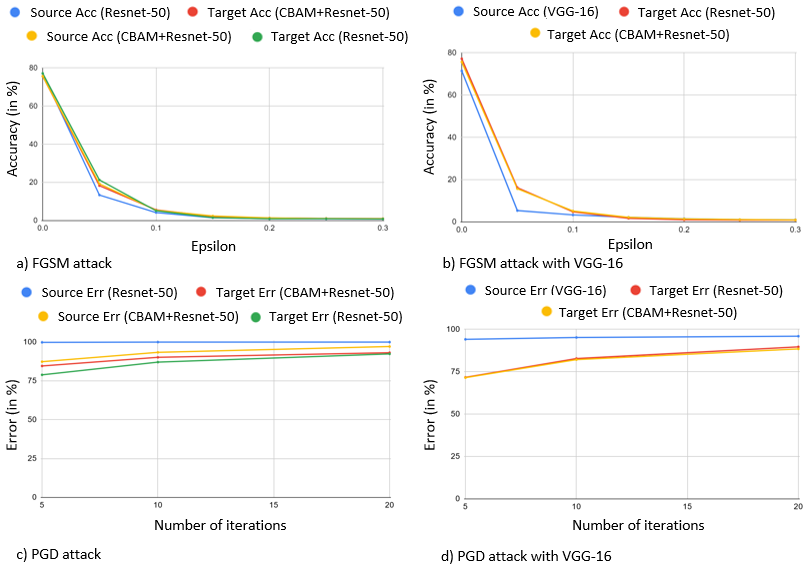}
    \caption{a) Transfer based FGSM, b) Transfer based FGSM with VGG-16 as source, c) Transfer based PGD and d) Transfer based PGD with VGG-16 as source for CIFAR-100 dataset.}
    \label{fig6}
\end{figure}

\subsubsection{CIFAR-100}
For the FGSM attack, Resnet-50 showed better transferability leading to a better attack on CBAM+Resnet-50 as compared to an attack on Resnet-50 by examples generated by CBAM+Resnet-50 (Fig.~\ref{fig6}(a)). Under the FGSM transfer attack with VGG-16 as the source, the attention and non-attention model both showed approximately equal accuracy under attack (Fig.~\ref{fig6}(b)). Under the transfer based PGD attack, Resnet-50 was comparatively more robust to attack against adversarial examples generated by CBAM+Resnet-50 (Fig.~\ref{fig6}(c)). Resnet-50 thus showed better transferability for the attack against attention based model. However, under transfer based PGD with VGG-16 as the source, CBAM+Resnet-50 shows better robustness as compared to Resnet-50 (Fig.~\ref{fig6}(d)).

\subsubsection{Fashion MNIST}
\begin{figure}[H]
    \centering
    \includegraphics[width=\linewidth]{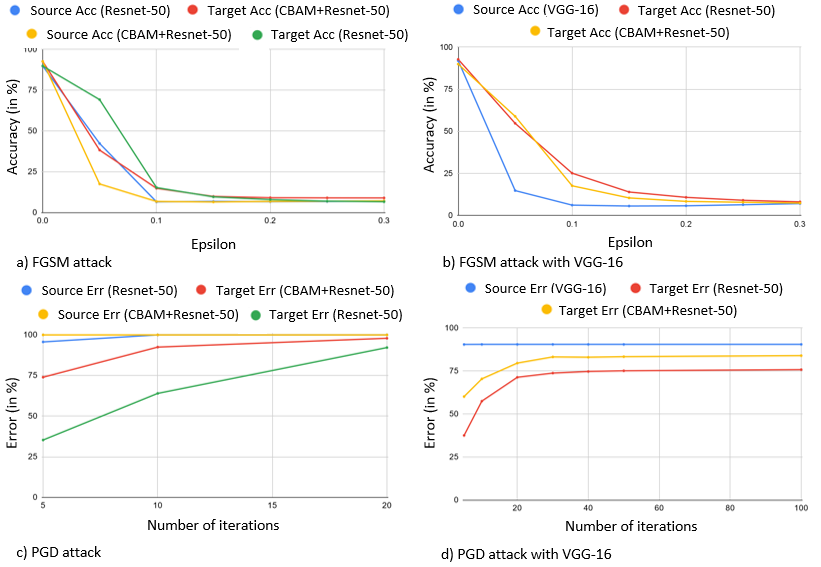}
    \caption{a) Transfer based FGSM, b) Transfer based FGSM with VGG-16 as source, c) Transfer based PGD and d) Transfer based PGD with VGG-16 as source for Fashion MNIST dataset.}
    \label{fig7}
\end{figure}

Resnet-50 is much more robust to transfer based FGSM attack from adversarial examples generated by CBAM+Resnet-50. Comparison between source and target accuracy in Fig.~\ref{fig7}(a) shows that Resnet-50 has good transferability since the accuracy of CBAM+Resnet-50 under attack differs from Resnet-50 as a source by a small margin. The transferability of CBAM+Resnet-50 under FGSM attack is much poorer which can be observed through the large difference in accuracy between CBAM+Resmet-50 (source) and Resnet-50 (target) (Fig.~\ref{fig7}(b)). Under the FGSM transfer based attack with VGG-16 as a source, there’s not much difference in the robustness of the non-attention and attention model. The attention based model (CBAM+Resnet-50) is slightly more robust for perturbation values greater than 0.5. The results of the transfer PGD attack in Fig 17 indicate that Resnet-50 is much more robust under examples generated by CBAM+Resnet-50 in comparison with the robustness of CBAM+Resnet-50 under adversarial examples by Resnet-50 (Fig.~\ref{fig7}(c)). A similar result is observed for PGD attack with VGG-16 as source where Resnet-50 shows better robustness against adversarial examples generated by VGG-16 (Fig.~\ref{fig7}(d)).

\section{Conclusion}
While it was expected that attention based models would be more robust under white box attacks and transfer based black box attack and would also show better transferability, the results obtained for CIFAR-10 and Fashion MNIST dataset strongly favour the robustness of the non-attention model i.e. Resnet-50. However, for the CIFAR-100 dataset, the attention based model i.e. CBAM+Resnet-50 was comparatively much more robust than the non-attention model Resnet-50 under white box attacks while the difference in results between the two models was very low in the transfer based black box attacks. These results suggest that the impact of attention on the robustness of image classification models could be dependent on the datasets used i.e. the number of classes in the datasets. Similarity in results of CIFAR-10 and Fashion MNIST and deviation of CIFAR-100 from these results could be due to fewer number of classes i.e. 10 classes in comparison with CIFAR-100 dataset which has 100 classes. The results obtained in this work is a step closer towards a generalized understanding of the impact of attention on the robustness of image classification models. Experiments with more datasets with different number of classes and models with other kinds of attention could be some directions for future work.

\section*{Acknowledgment}
	We thank our institute, Indian Institute of Information Technology Allahabad (IIITA), India and Big Data Analytics (BDA) lab for allocating the centralised computing facility and other necessary resources to perform this research. We extend our thanks to our colleagues for their valuable guidance and suggestions.

\bibliographystyle{IEEEtran}
\bibliography{reference}

\end{document}